\documentclass{article}

\usepackage{microtype}
\usepackage{graphicx}
\usepackage{subcaption}
\usepackage{booktabs} 
\usepackage{enumitem}
\usepackage{makecell}
\usepackage{multirow}

\usepackage{hyperref}

\usepackage[preprint]{icml2026}

\usepackage{amsmath}
\usepackage{amssymb}
\usepackage{mathtools}
\usepackage{amsthm}
\usepackage[table,xcdraw]{xcolor}  
\usepackage{colortbl}

\usepackage[capitalize,noabbrev]{cleveref}

\theoremstyle{plain}

\theoremstyle{definition}

\theoremstyle{remark}

\usepackage[textsize=tiny]{todonotes}

\icmltitlerunning{TextME: Bridging Unseen Modalities Through Text Descriptions}

\begin{document}

\twocolumn[
  \icmltitle{TextME: Bridging Unseen Modalities Through Text Descriptions}

  \icmlsetsymbol{equal}{*}

  \begin{icmlauthorlist}
    \icmlauthor{Soyeon Hong}{ajou}
    \icmlauthor{Jinchan Kim}{ajou}
    \icmlauthor{Jaegook You}{ajou}
    \icmlauthor{Seungtaek Choi}{hufs}
    \icmlauthor{Suha Kwak}{postech}
    \icmlauthor{Hyunsouk Cho}{ajou-s}
  \end{icmlauthorlist}

  \icmlaffiliation{ajou}{Department of Artificial Intelligence, Ajou University, Suwon, South Korea}
  \icmlaffiliation{hufs}{Division of Language \& AI, Hankuk University of Foreign Studies, Seoul, Korea}
  \icmlaffiliation{postech}{Graduate School of AI, POSTECH, Pohang, Korea}
  \icmlaffiliation{ajou-s}{Department of Software, Ajou University, Suwon, South Korea}

  \icmlcorrespondingauthor{Hyunsouk Cho}{hyunsouk@ajou.ac.kr}

  \icmlkeywords{Machine Learning, Multimodal Learning, Large Language Model}

  \vskip 0.3in
]

\printAffiliationsAndNotice{}  

\newcommand{\eg}{e.g.}
\newcommand{\ie}{i.e.}

\newcommand{\ours}{\textbf{TextME}}
\newcommand{\fours}{Text-only Modality Expansion (\textbf{TextME})}
\newcommand{\oursuppaer}{\textbf{Ours$_{upper\text{-}bound}$}}

\newcommand{\task}{Modality Expansion}

\newcommand{\todoc}[2]{{\textcolor{#1}{\textbf{#2}}}}
\newcommand{\todoorange}[1]{\todoc{orange}{\textbf{[[#1]]}}}
\newcommand{\hist}[1]{\todoorange{hist: #1}}

\newcommand{\romannum}[1]{\romannumeral #1}
\newcommand{\com}{\textcolor{red}}
\newcommand{\revise}{\textcolor{blue}}
\begin{abstract}
Expanding multimodal representations to novel modalities is constrained by reliance on large-scale paired datasets (\eg, text–image, text–audio, text–3D, text–molecule), which are costly and often infeasible in domains requiring expert annotation such as medical imaging and molecular analysis. We introduce \ours, the first text-only modality expansion framework, to the best of our knowledge, projecting diverse modalities into LLM embedding space as a unified anchor. Our approach exploits the geometric structure of pretrained contrastive encoders to enable zero-shot cross-modal transfer using only text descriptions, without paired supervision. We empirically validate that such consistent modality gaps exist across image, video, audio, 3D, X-ray, and molecular domains, demonstrating that text-only training can preserve substantial performance of pretrained encoders. We further show that our framework enables emergent cross-modal retrieval between modality pairs not explicitly aligned during training (\eg, audio-to-image, 3D-to-image). These results establish text-only training as a practical alternative to paired supervision for modality expansion. Our code is publicly available at \url{https://github.com/SoyeonHH/TextME}.
\end{abstract}
\section{Introduction}

\begin{figure}
    \centering
    \includegraphics[width=1\linewidth]{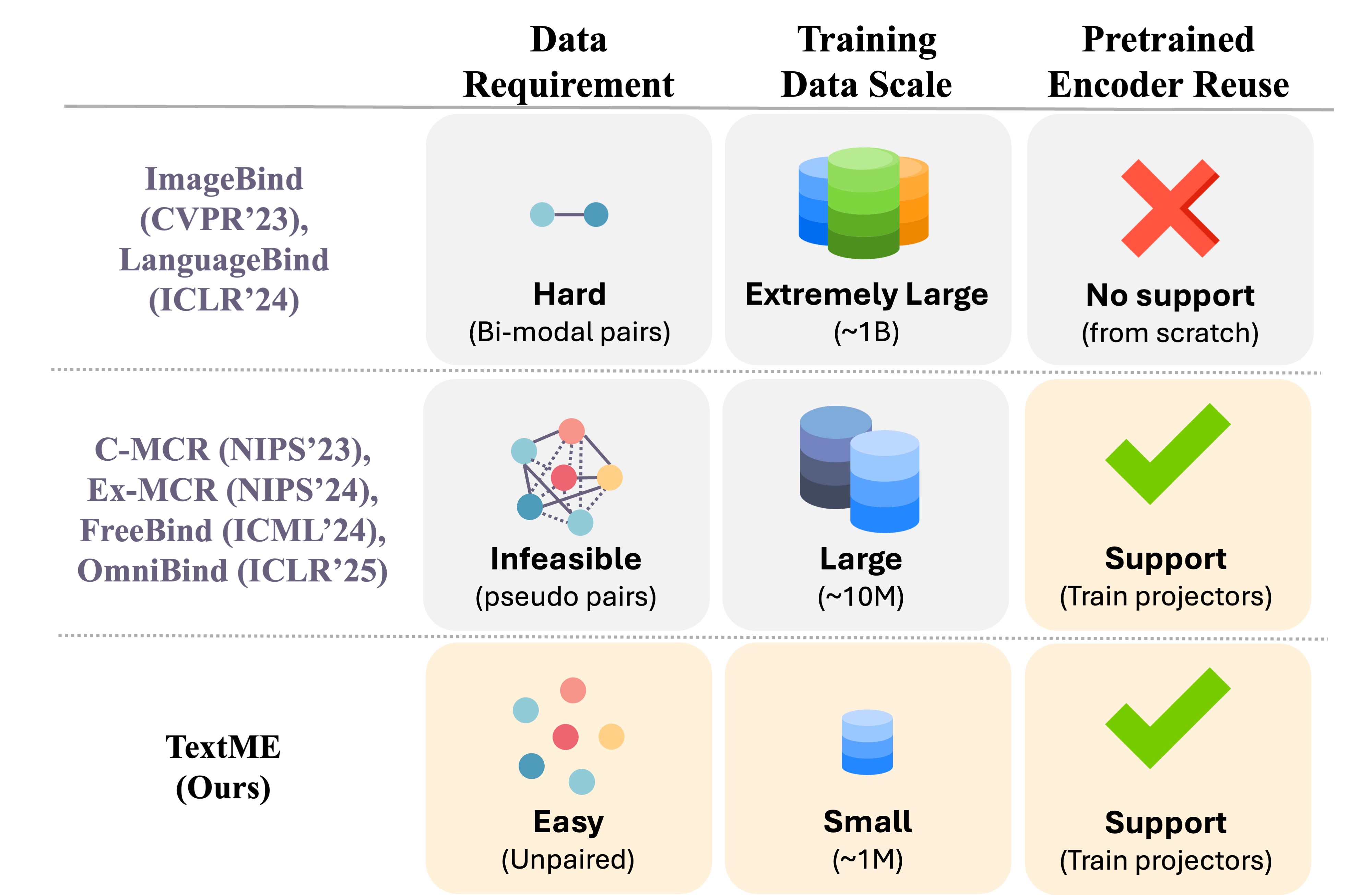}
    \caption{\textbf{Comparison of modality expansion approaches}. Unlike prior methods that require large-scale paired data or pseudo-pair construction through overlapping encoders, \ours\ achieves modality expansion using only unpaired text descriptions while reusing pretrained encoders.}
    \label{fig:teaser}
    \vspace{-10pt}
\end{figure}

Modality expansion, which aligns heterogeneous data modalities into a unified embedding space, has emerged as a core challenge in multimodal representation learning \citep{baltruvsaitis2018multimodal, manzoor2023multimodality, liang2024foundations, yuan2025survey}. Recent approaches leverage large-scale paired datasets to project diverse modalities—such as images, audio, and 3D point clouds—into shared semantic spaces where equivalent content maintains proximity \citep{zhang2023meta, han2023imagebind, zhu2023languagebind, lyu2024unibind, guo2023point}. While text–image and text–audio corpora have enabled remarkable progress in vision–language \citep{radford2021learning, jia2021scaling} and audio–language modeling \citep{wu2023large, manco2022contrastive}, extending this paradigm to specialized domains proves prohibitively expensive or infeasible. Medical imaging requires costly expert annotations while navigating privacy constraints \citep{wang2025cxpmrg, ziller2021medical}, molecular analysis demands complex domain-specific representations \citep{xiao2024molbind}, and 3D modeling necessitates labor-intensive curation \citep{deitke2023objaverse}. Consequently, the scalability of modality expansion remains fundamentally limited by the availability of paired supervision.

Recent methods reduce computational costs by reusing pretrained encoders through lightweight projection networks \citep{wang2023connecting, zhang2024extending, wang2024freebind, wang2024omnibind}, yet they still require constructing semantically aligned pseudo pairs across all target modalities through overlapping encoders. Meanwhile, prior work has revealed that contrastive encoders exhibit a consistent modality gap—a systematic offset between text and modality embeddings—that can enable cross-modal transfer via simple geometric operations \citep{liang2022mind, zhang2023diagnosing, zhang2024connect}. However, these studies have primarily focused on analyzing the gap in vision-language models or mitigating the gap within paired-data settings; whether this geometric property can be exploited to eliminate the need for paired supervision altogether remains unexplored. In addition, there is no guidance on which modalities are amenable to effective alignment and which are not.

In this work, we demonstrate that the modality gap can enable modality expansion without paired supervision. We propose \ours, a framework that projects modality-specific embeddings into LLM embedding space as a unified semantic anchor by applying precomputed offset corrections derived from the gap structure.
As illustrated in Figure~\ref{fig:teaser}, unlike prior methods that require large-scale bi-modal pairs or pseudo-pair construction through overlapping encoders, our proposed framework achieves modality expansion using only unpaired text descriptions---with substantially reduced data requirements while fully leveraging pretrained encoders through lightweight projectors.

We evaluate the framework across six diverse modalities—image, video, audio, 3D point clouds, X-ray, and molecules—on both cross-modal retrieval and zero-shot classification tasks. Our experiments demonstrate that \ours\ achieves competitive performance relative to paired-data methods and, notably, enables emergent cross-modal capabilities between modality pairs never observed during training, such as audio-to-3D and molecule-to-image retrieval. These results suggest that text modality can create meaningful semantic bridges across arbitrary modalities without explicit cross-modal supervision. To better understand the variation in performance across modalities, we further analyze the geometric properties of each encoder and find that the consistency of gap-content orthogonality correlates with downstream performance, providing insight into when text-only expansion is most effective.

Our contribution is three-fold:
\begin{itemize}[itemsep=4pt, topsep=2pt, parsep=4pt]
    \item We propose \ours, a text-only modality expansion framework that exploits modality gap geometry to learn cross-modal projections using only text descriptions, eliminating the need for paired multimodal supervision during training.
    
    \item We investigate LLM embedding space as a unified anchor for modality expansion and compare it against multimodal encoder representations, analyzing their varying effectiveness across tasks and modalities.
    
    \item We empirically validate the framework across six diverse modalities, demonstrating competitive performance on retrieval and classification tasks and identifying encoder characteristics that predict when text-only expansion is most effective.
\end{itemize}

\section{Preliminaries}
\label{sec:preliminary}

\subsection{Problem Formulation}

Modality expansion aims to integrate pretrained modality-specific encoders into a unified semantic space where similar concepts maintain proximity regardless of their source modality. Let $\mathcal{M} = \{m_1, \ldots, m_k\}$ denote a set of target modalities to be aligned. For each modality $m \in \mathcal{M}$, a pretrained contrastive encoder consists of a text branch $E_m^{\text{text}}: \mathcal{T} \rightarrow \mathbb{R}^{d_m}$ and a modal branch $E_m^{\text{modal}}: \mathcal{X}_m \rightarrow \mathbb{R}^{d_m}$, where $\mathcal{T}$ is the space of text descriptions and $\mathcal{X}_m$ is the input space for modality $m$. Our objective is to learn projection networks $P_m: \mathbb{R}^{d_m} \rightarrow \mathbb{R}^{d_h}$ that map modal embeddings into a shared $d_h$-dimensional anchor space.


Existing methods require instance-level paired data $\{(x_i, t_i)\}_{i=1}^{N}$ of modal inputs $x_i \in \mathcal{X}_m$ and text descriptions $t_i \in \mathcal{T}$ to train cross-modal projections~\citep{han2023imagebind,zhu2023languagebind,lyu2024unibind}. Recent approaches connect multiple pretrained encoders via overlapping modalities: given encoders pretrained on modality pairs $(\mathcal{A}, \mathcal{B})$ and $(\mathcal{B}, \mathcal{C})$, they leverage data from the shared modality $\mathcal{B}$ to align encoder spaces, enabling transfer to non-overlapping pairs $(\mathcal{A}, \mathcal{C})$~\citep{wang2023connecting, zhang2024extending,wang2024freebind,wang2024omnibind}. This requires modality overlap across all target encoders. In this work, we consider a more practical scenario: learning projection networks independently for each modality using only unpaired text descriptions $\{t_i\}_{i=1}^{N}$, without requiring cross-encoder alignment or access to target modality samples.

\begin{figure*}
    \centering
    \includegraphics[width=1\linewidth]{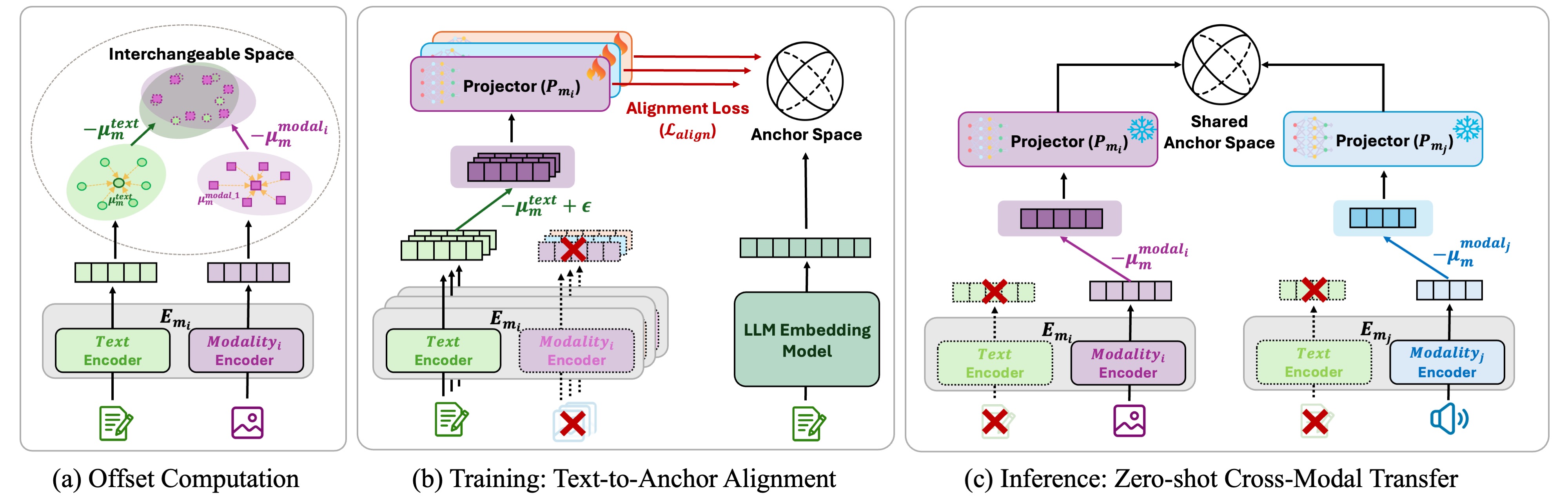}
    \caption{\textbf{Overview of the TextME pipeline}. (a) Offset computation estimates modality-specific centroids from unpaired samples, creating an interchangeable space where centered text and modal embeddings become functionally equivalent. (b) During training, projection networks are learned by aligning centered text embeddings with a unified LLM anchor space, requiring only text descriptions. (c) At inference, centering modal embeddings with the precomputed offset enables zero-shot cross-modal transfer without paired supervision.}
    \label{fig:framework}
\end{figure*}

\subsection{Modality Gap and Interchangeable Space}

Prior work has shown that contrastive encoders trained with objectives such as InfoNCE exhibit a systematic offset between text and modal embedding spaces \citep{liang2022mind,zhang2023diagnosing,zhang2024connect}. For each encoder $E_m$, the modality gap is characterized by the difference between the centroids of modal and text embeddings:
\begin{equation}
    \Delta_m = \mu_m^{\text{modal}} - \mu_m^{\text{text}},
\end{equation}
where $\mu_m^{\text{modal}} = \mathbb{E}[E_m^{\text{modal}}(x)]$ and $\mu_m^{\text{text}} = \mathbb{E}[E_m^{\text{text}}(t)]$ denote the expected embeddings over their respective distributions. This gap presents a fundamental challenge for text-only training, as projection networks learned from text embeddings cannot directly transfer to modal embeddings that occupy a different region of the space.

\paragraph{Interchangeable Space via Centering.}
A key observation from \citet{zhang2024connect} is that this challenge can be addressed through independent centering operations. Consider a semantically matched pair $(t, x)$ with embeddings $e_t = E_m^{\text{text}}(t)$ and $e_x = E_m^{\text{modal}}(x)$. Although these embeddings differ due to the modality gap, subtracting their respective centroids yields centered embeddings
\begin{equation}
    \hat{e}_t = e_t - \mu_m^{\text{text}}, \quad \hat{e}_x = e_x - \mu_m^{\text{modal}},
\end{equation}
that satisfy $\hat{e}_t \approx \hat{e}_x$ for semantically corresponding pairs. That is, centering removes the modality-specific bias while preserving the shared semantic content, creating an \emph{interchangeable space} where text and modal embeddings become functionally equivalent~\citep{an2025i0t}. This property enables projection networks trained on centered text embeddings to generalize to centered modal embeddings at inference time, forming the basis of our text-only training approach.

\section{\ours: Text-only Modality Expansion}

We present \ours, a framework that enables modality expansion using only text descriptions by exploiting the geometric properties of pretrained contrastive encoders. Figure~\ref{fig:framework} illustrates the overall pipeline.

\subsection{Overview}

The key insight of \ours\ is that the interchangeable space described in Section~\ref{sec:preliminary} allows projection networks trained on centered text embeddings to generalize to centered modal embeddings at inference time. Our framework operates in two phases. During training, we precompute modality-specific centroids and train lightweight projection networks to map centered text embeddings into a shared anchor space. At inference, we apply the same centering operation to modal embeddings before projection, enabling zero-shot cross-modal transfer without having observed any modal samples during training.

\subsection{Offset Computation}

As established in Section~\ref{sec:preliminary}, creating an interchangeable space requires estimating the centroids $\mu_m^{\text{text}}$ and $\mu_m^{\text{modal}}$ for each modality. We compute these centroids from representative samples:
\begin{equation}
\mu_m^{\text{text}} = \frac{1}{N}\sum_{i=1}^{N} E_m^{\text{text}}(t_i), \quad \mu_m^{\text{modal}} = \frac{1}{M}\sum_{j=1}^{M} E_m^{\text{modal}}(x_j),
\end{equation}
where $\{t_i\}_{i=1}^N \subset \mathcal{T}$ and $\{x_j\}_{j=1}^M \subset \mathcal{X}_m$ are sampled independently from text and modal distributions. Unlike projection training, these samples need not be instance-level paired—only representative coverage of each distribution is required for accurate centroid estimation.

Importantly, accurate centroid estimation requires only a small number of samples. In our experiments, we find that 5K samples suffice for stable estimation across all evaluated modalities, representing less than 5\% of typical paired training requirements~\citep{zhu2023languagebind, zhang2024extending}. The centroids are precomputed once and remain fixed throughout training.

\subsection{Text-to-Anchor Alignment}

Given the precomputed offsets, we train projection networks using only text descriptions from the target domain. For each modality $m$, a projection network $P_m: \mathbb{R}^{d_m} \rightarrow \mathbb{R}^{d_h}$ maps centered text embeddings into a shared anchor space.

\begin{figure}
    \centering
    \includegraphics[width=0.8\linewidth]{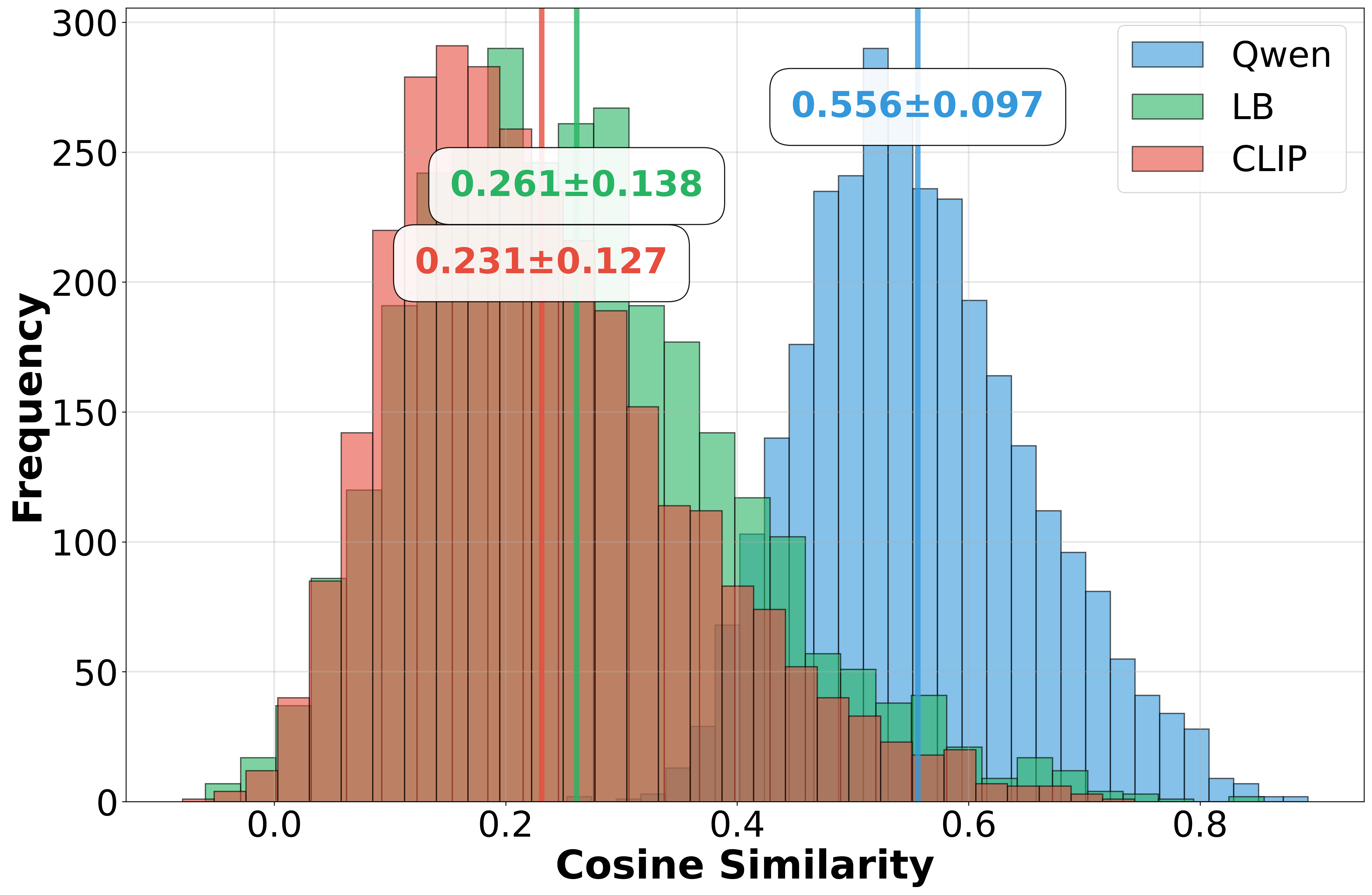}
    \caption{\textbf{Semantic anchoring comparison.} LLM embeddings and multimodal encoders are compared on 3K semantically equivalent cross-modal description pairs. LLM embeddings exhibit clearer separation between matched and unmatched pairs, demonstrating superior cross-domain alignment capability.}
    \label{fig:anchor}
\end{figure}

\paragraph{Anchor Space Selection.}
We adopt LLM embedding space as our unified anchor rather than multimodal text encoders. While multimodal encoders such as CLIP are optimized for cross-modal matching, LLMs trained on large-scale text corpora capture richer semantic relationships that generalize across diverse domains. To assess cross-domain alignment capabilities, we analyze 3K audio-image caption pairs from FlickrNet~\citep{senocak2018learning}, where we generated linguistically distinct but semantically equivalent descriptions using the Gemini API~\citep{gemini2024api}—for instance, an image caption ``a red sports car speeding on highway'' is paired with its audio equivalent ``loud engine roar with wind rushing past.'' As shown in Figure~\ref{fig:anchor}, LLM embeddings (\ie, Qwen) exhibit clearer separation between semantically equivalent and unrelated pairs (0.56 vs.\ 0.23--0.26 mean cosine similarity) compared to multimodal encoders, suggesting better suitability for bridging heterogeneous descriptions. This advantage is further corroborated by semantic textual similarity benchmarks, where LLM embeddings achieve Spearman correlations of 85--90 compared to 67--68 for multimodal encoders (see Appendix~\ref{app:sts} for details). Based on these findings, we adopt Qwen3-Embedding~\citep{zhang2025qwen3} as our default anchor space.

\paragraph{Training Objective.}
Given text descriptions $\mathcal{D}_{\text{text}} = \{t_i\}_{i=1}^{N}$ from the target modality domain, we train the projection network by aligning centered text embeddings with their corresponding LLM embeddings:
\begin{equation}
    \mathcal{L}_{\text{align}} = -\frac{1}{B} \sum_{i=1}^{B} 
    \log \frac{\exp(\text{sim}(z_i, z'_i) / \tau)}
    {\sum_{j \in \mathcal{N}_i \cup \{i\}} \exp(\text{sim}(z_i, z'_j) / \tau)}
\end{equation}
where $z_i = P_m(\hat{e}_{t_i})$ is the projected centered text embedding with $\hat{e}_{t_i} = E_m^{\text{text}}(t_i) - \mu_m^{\text{text}}$, $z'_i = E_{\text{LLM}}(t_i)$ is the corresponding LLM embedding, and $\mathcal{N}_i$ contains hard negatives. Following recent language embedding models~\citep{lee2024nv, moreira2024nv, rosch2024enhancing}, we employ hard negative mining to focus training on challenging examples near the decision boundary, improving the discriminative quality of learned projections.

\subsection{Inference}

At inference time, \ours\ enables zero-shot cross-modal transfer by mapping modal embeddings into the interchangeable space. For a non-text input $x$ from modality $m$, the final embedding is computed as:
\begin{equation}
    e_{\text{final}} = P_m(\hat{e}_x) = P_m(E_m^{\text{modal}}(x) - \mu_m^{\text{modal}}).
\end{equation}
The centering operation transforms the modal embedding into the interchangeable space where text embeddings reside during training. As established in Section~\ref{sec:preliminary}, centering preserves semantic relationships while removing modality-specific bias, allowing the text-trained projection network $P_m$ to process modal inputs without modification. The final embeddings are thus aligned within the unified LLM anchor space alongside text representations, enabling direct cross-modal retrieval and zero-shot classification.

\section{Experiments}
\label{sec:experiments}

\begin{table*}[t]
\caption{\textbf{Zero-shot performance across all evaluation benchmarks.} \textbf{PPR}: Performance Preservation Ratio (\%) relative to pretrained encoders. $\times$ indicates unavailable results due to missing official implementations or incompatible evaluation protocols. Bold indicates best among unpaired methods. $\dagger$Our reproduction.}
\label{tab:main}
\centering
\scriptsize
\setlength{\tabcolsep}{4pt}
\begin{tabular}{l cccccccc ccccc cc l}
\toprule
& \multicolumn{8}{c}{\textbf{Text$\to$X Retrieval}} 
& \multicolumn{5}{c}{\textbf{Classification}} 
& \multicolumn{2}{c}{\textbf{Emergent X$\to$X}} 
& \\
\cmidrule(lr){2-9} \cmidrule(lr){10-14} \cmidrule(lr){15-16}
& \multicolumn{2}{c}{Image} & \multicolumn{3}{c}{Video} & \multicolumn{2}{c}{Audio} & Mol. 
& \multicolumn{2}{c}{Audio} & \multicolumn{2}{c}{3D} & X-ray
& A$\to$I & 3D$\to$I 
& Data \\
& COCO & Flkr. & MSR. & MSVD & DiDe. & ACaps. & Clo. & Drug. 
& ASet. & ESC & MN40. & Scan. & RSNA 
& Flkr. & Obja. 
& Requirements \\
\midrule
\rowcolor{gray!10}
Pretrained & 48.29 & 77.70 & 37.00 & 51.06 & 31.27 & 22.47 & 16.90 & 79.19 
           & 9.32 & 85.20 & 67.75 & 42.21 & 52.64 
           & $\times$ & $\times$ 
           & \\
\midrule
\multicolumn{17}{l}{\textit{Paired-data methods}} \\
LanguageBind & 44.53 & 73.42 & 45.30 & 65.22 & 36.85 & 12.42 & 11.32 & $\times$ 
             & 18.33 & 94.00 & $\times$ & $\times$ & $\times$ 
             & 1.52 & $\times$ 
             & 10M pairs \\
Ex-MCR & 40.24 & 71.89 & $\times$ & $\times$ & $\times$ & 19.07 & 7.01 & $\times$ 
       & 6.67 & 71.20 & 66.53 & 40.31 & $\times$ 
       & 1.57 & 5.67 
       & 1M pairs$^*$ \\
\midrule
\multicolumn{17}{l}{\textit{Unpaired-data methods}} \\
Na\"ive & 0.01 & 0.04 & 0.00 & 0.00 & 0.00 & 0.02 & 0.04 & 10.17 
        & 1.14 & 2.90 & 0.81 & 3.32 & 26.36 
        & 0.02 & 0.00 
        & 0 \\
COX$^\dagger$ & 0.02 & 0.20 & 5.10 & 0.00 & 0.10 & 0.08 & 0.11 & 7.63 
              & 1.26 & 2.00 & 4.05 & 2.84 & 22.53 
              & 0.02 & 0.00 
              & 10K labels \\
\rowcolor{cyan!8}
\textbf{TextME} & \textbf{28.63} & \textbf{51.66} & \textbf{26.40} & \textbf{45.82} & \textbf{24.10} & \textbf{15.35} & \textbf{7.81} & \textbf{34.75} 
                & \textbf{5.80} & \textbf{77.25} & \textbf{70.86} & \textbf{42.15} & \textbf{46.59} 
                & \textbf{1.06} & \textbf{10.27} 
                & \textbf{100K text} \\
\midrule
\textbf{PPR(\%)} 
& \cellcolor{yellow!25}59.3 
& \cellcolor{yellow!32}66.5 
& \cellcolor{yellow!40}71.4 
& \cellcolor{yellow!58}89.7 
& \cellcolor{yellow!45}77.1 
& \cellcolor{yellow!35}68.3 
& \cellcolor{yellow!18}46.2 
& \cellcolor{yellow!15}43.9 
& \cellcolor{yellow!28}62.2 
& \cellcolor{yellow!60}90.7 
& \cellcolor{yellow!75}104.6 
& \cellcolor{yellow!70}99.9 
& \cellcolor{yellow!55}88.5 
& $\times$ 
& $\times$ 
& \\
\bottomrule
\multicolumn{17}{l}{\footnotesize $^*$Indirect: uses overlapping modality from existing MCR spaces. TextME requires \textbf{zero paired data} and \textbf{zero labeled target data}.}
\end{tabular}
\end{table*}

\begin{figure*}[t]
    \centering
    \includegraphics[width=0.75\linewidth]{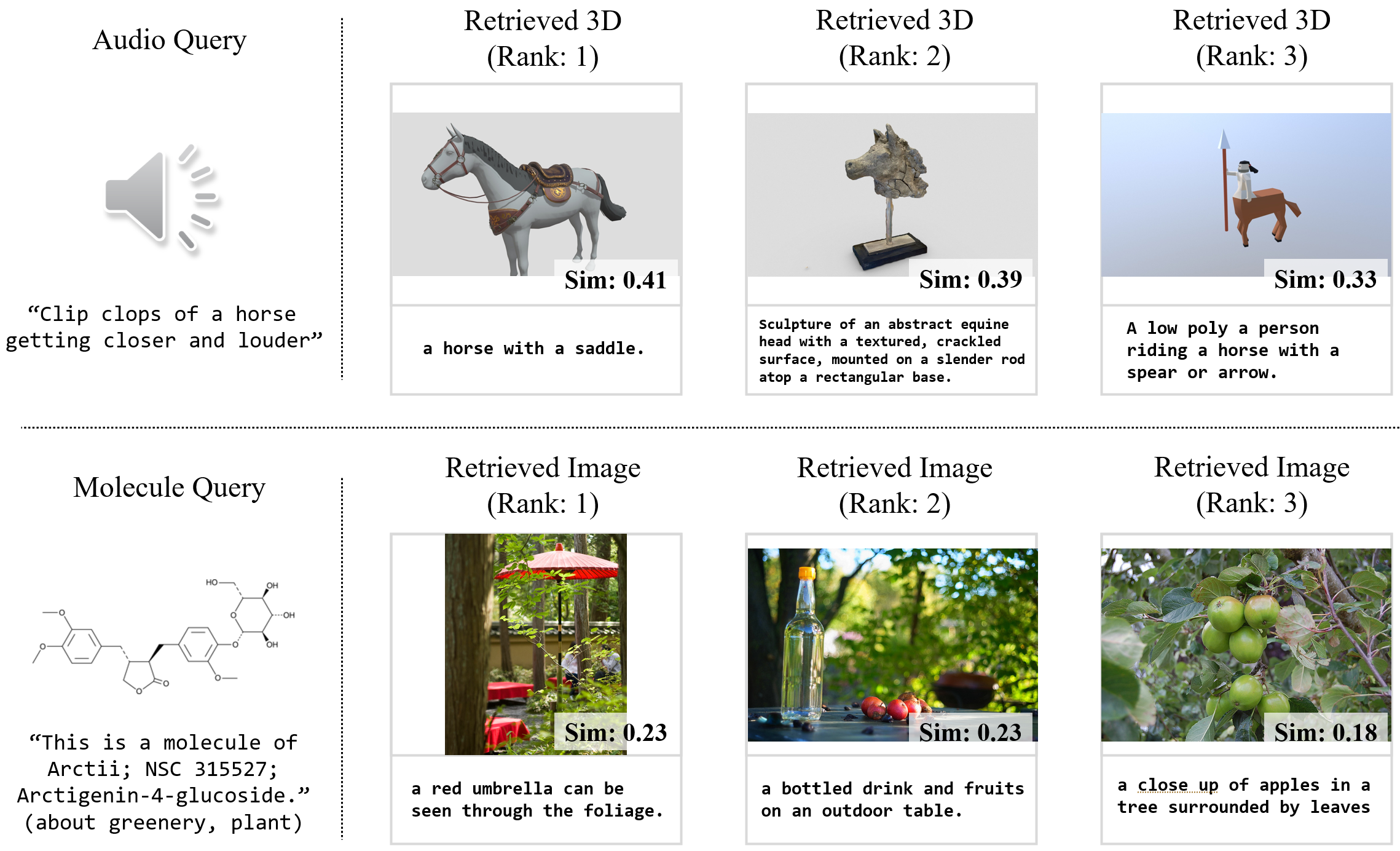}
    \caption{\textbf{Emergent cross-modal retrieval without paired supervision.} Audio queries retrieve semantically related 3D objects (top), and molecular structures retrieve contextually appropriate images (bottom). These modality pairs were never seen during training, demonstrating that text-anchored alignment creates semantic bridges across arbitrary modalities.}
    \label{fig:emergent}
\end{figure*}

We evaluate \ours\ on cross-modal retrieval and zero-shot classification across six modalities: image, video, audio, 3D, X-ray, and molecules. Our experiments address three key questions: (1) whether text-only training can achieve competitive performance relative to paired-data methods and pretrained encoders (Section~\ref{sec:main_results}); (2) what geometric properties predict success or failure across different modalities (Section~\ref{sec:analysis}); and (3) how different design choices including anchor space selection and offset correction influence overall performance (Section~\ref{sec:ablation}).

\subsection{Experimental Setup}

\paragraph{Modalities and Encoders.}
We adopt LanguageBind~\citep{zhu2023languagebind} as the text encoder for all text-to-modal retrieval and zero-shot classification tasks. For modal encoders, we adopt CLIP~\citep{radford2021learning} for image, ViCLIP~\citep{wang2023internvid} for video, CLAP~\citep{elizalde2023clap} for audio, Uni3D~\citep{zhou2023uni3d} for 3D, CXR-CLIP~\citep{you2023cxr} for X-ray, and MoleculeSTM~\citep{liu2023multi} for molecule. Each encoder pair is independently projected into the shared LLM anchor space for cross-modal matching. We sample $100K$ text descriptions per modality for projection training, with offset computation on $5K$ samples.

\paragraph{Baselines.}
We compare against three categories of methods. First, we report the performance of the original \textit{pretrained encoders} as reference points. Second, for \textit{paired-data approaches}, we include LanguageBind~\citep{zhu2023languagebind} and Ex-MCR~\citep{zhang2024extending}, both of which perform modality expansion using fully-paired multimodal data. Third, for \textit{unpaired-data methods}, we compare with COX$^\dagger$~\citep{huang2025towards}, which learns target modality representations from scratch without instance-level pairing but requires substantial target modality data and classification labels. We also include a Na\"ive baseline that simply aligns embedding dimensions via PCA without any learned projection. Unlike COX, \ours\ requires no target modality data during training.

\paragraph{Evaluation.}
We evaluate on three task categories: Text$\to$X retrieval, emergent cross-modal retrieval between unseen modality pairs, and zero-shot classification. 
Table~\ref{tab:main} reports results on representative benchmarks per modality, selected based on prevalence in prior work~\citep{zhu2023languagebind, zhang2024extending}.
We report Recall@$k$ (R@$k$) for retrieval, MRR@$k$ for molecule retrieval following~\citet{liu2023multi}, and Top-$k$ accuracy for classification.
We define \emph{Performance Preservation Ratio} (PPR) as the percentage of pretrained encoder performance retained by our method: $\text{PPR} = (\text{\ours\ score} / \text{Pretrained score}) \times 100\%$.
Complete results across all benchmarks appear in Appendix~\ref{app:extended_results}.

\vspace{-5pt}
\subsection{Does Text-Only Training Preserve Pretrained Performance?}
\label{sec:main_results}

Table~\ref{tab:main} reports performance across all evaluation tasks. \ours\ achieves an average of 74.5\% PPR across all tasks, with classification at 89.2\% consistently outperforming retrieval at 65.3\%, suggesting that offset-based alignment preserves categorical boundaries more effectively than fine-grained similarity structure. Among unpaired baselines, COX~\citep{huang2025towards} yields substantially lower performance, as it requires a pretrained classifier on labeled target data. Since official implementations are not publicly available, we trained this classifier from scratch on evaluation data. In contrast, our framework eliminates the need for labeled target data and paired supervision, thereby enabling direct generalization to novel modalities.

\paragraph{Zero-Shot Retrieval and Classification.}
Across both task categories, \ours\ substantially outperforms unpaired baselines and achieves comparable results to paired-data methods. As reported in the \textit{Data Requirements} column of Table~\ref{tab:main}, this performance is achieved using only 100K text descriptions, compared to 1--10M paired samples required by methods such as LanguageBind and Ex-MCR. This reduction of over two orders of magnitude in supervision requirements makes modality expansion practical for specialized domains where paired annotation is prohibitively expensive. In terms of task-specific patterns, classification consistently demonstrates higher preservation than retrieval, as categorical discrimination requires only well-separated decision boundaries whereas retrieval demands fine-grained instance-level similarity that is more sensitive to distortions introduced by offset correction. Notably, 3D zero-shot classification surpasses pretrained Uni3D with 104.6\% PPR on ModelNet40, while retrieval preservation varies substantially across modalities. We examine the factors underlying these variations in Section~\ref{sec:analysis}.

\paragraph{Emergent Cross-Modal Capabilities.}
The unified anchor space additionally enables retrieval between modality pairs not explicitly aligned during training. As reported in the Emergent X$\rightarrow$X columns of Table~\ref{tab:main}, \ours\ outperforms Ex-MCR on 3D$\rightarrow$Image despite the latter requiring paired supervision, and achieves comparable performance to paired-data methods on Audio$\rightarrow$Image. To qualitatively examine whether the learned representations enable retrieval between modality pairs without any paired annotations, we conduct cross-modal retrieval experiments using independently collected datasets. Specifically, we sample instances from each modality and perform retrieval across disjoint modality pairs such as Audio$\rightarrow$3D and Molecule$\rightarrow$Image. Figure~\ref{fig:emergent} presents representative results, demonstrating that audio queries retrieve semantically coherent 3D models and molecular queries retrieve contextually relevant images. These findings suggest that text-anchored alignment establishes implicit semantic correspondences without explicit cross-modal supervision. 

\begin{table}[t]
    \centering
    \caption{\textbf{Geometric properties of contrastive encoders.} (i) intra-modal independence, (ii) gap consistency, (iii) bounded deviation, (iv) gap-content orthogonality. $\dagger$: weaker satisfaction ($>$0.1 for (i), $<$0.96 for (ii)).}
    \label{tab:geometry}
    \small
    \setlength{\tabcolsep}{2pt}
    \begin{tabular}{llcccc}
        \toprule
        \textbf{Encoder} & \textbf{Mod.} & \textbf{(i)$\downarrow$} & \textbf{(ii)$\uparrow$} & \textbf{(iii)$\downarrow$} & \textbf{(iv)$\downarrow$} \\
        \midrule
        CLIP        & Image    & .28$^\dagger${\tiny$\pm$.11} & .97{\tiny$\pm$.00} & .00{\tiny$\pm$.00} & .00{\tiny$\pm$.11} \\
        ViCLIP      & Video    & .18$^\dagger${\tiny$\pm$.11} & .98{\tiny$\pm$.00} & .00{\tiny$\pm$.00} & .00{\tiny$\pm$.06} \\
        CLAP        & Audio    & .12$^\dagger${\tiny$\pm$.18} & .97{\tiny$\pm$.01} & .00{\tiny$\pm$.00} & .00{\tiny$\pm$.15} \\
        Uni3D       & 3D       & .07{\tiny$\pm$.06}           & .96{\tiny$\pm$.00} & .00{\tiny$\pm$.00} & .00{\tiny$\pm$.04} \\
        CXR-CLIP    & X-ray    & .37$^\dagger${\tiny$\pm$.13} & .99{\tiny$\pm$.00} & .00{\tiny$\pm$.00} & .00{\tiny$\pm$.06} \\
        MoleculeSTM & Molecule & .01{\tiny$\pm$.19}           & .78$^\dagger${\tiny$\pm$.05} & .00{\tiny$\pm$.00} & .00{\tiny$\pm$.18} \\
        \bottomrule
    \end{tabular}
\end{table}

\subsection{When Does Text-Only Expansion Succeed?}
\label{sec:analysis}

We hypothesize that the effectiveness of text-only expansion depends on how well pretrained encoders satisfy the geometric properties underlying modality gap alignment. The results above support this view, revealing substantial variation in performance preservation across modalities—ranging from over 100\% for 3D classification to approximately 42\% for Molecule retrieval. To investigate this relationship, we measure the geometric characteristics of each encoder and examine their correlation with downstream performance.

\begin{figure}[t]
    \centering
    \includegraphics[width=0.95\columnwidth]{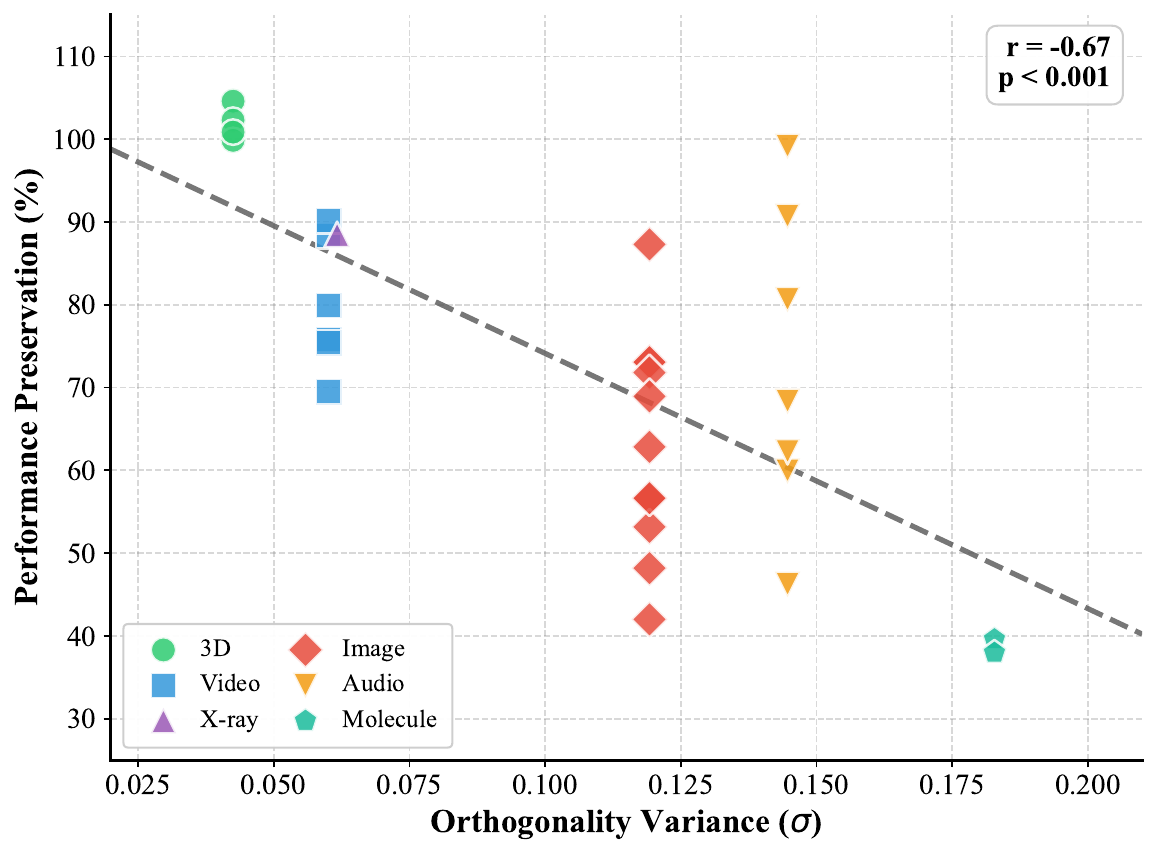}
    \caption{\textbf{Orthogonality variance vs.\ performance preservation.} Each point represents a single evaluation metric from six modalities, with tasks sharing the same encoder aligned vertically at identical variance values. Lower variance in gap-content orthogonality corresponds to higher downstream performance.}
    \label{fig:correlation}
\end{figure}

\paragraph{Geometric Properties.}
Following prior work~\citep{zhang2024connect}, we measure four geometric properties for each encoder using 5K paired samples:
(i) \textbf{Intra-modal independence}: $\mathbb{E}[\cos(\hat{e}, \hat{\mu}_m)]$, measuring whether embeddings are statistically independent from the modality centroid;
(ii) \textbf{Gap consistency}: $\cos(\Delta^{(k)}_m, \Delta_m)$, measuring whether instance-level offsets $\Delta^{(k)}_m = e^{(k)}_{\text{modal}} - e^{(k)}_{\text{text}}$ align directionally with the group-level offset $\Delta_m$;
(iii) \textbf{Bounded deviation}: $\text{std}(\epsilon_k)$ where $\epsilon_k = \Delta^{(k)}_m - \Delta_m$, measuring the variance of instance-level offsets around the mean;
(iv) \textbf{Gap-content orthogonality}: $|\cos(\Delta_m, r^{(p,q)})|$ where $r^{(p,q)} = e_p - e_q$, measuring whether the modality gap is independent of intra-modal semantic variations.
Properties (i) and (ii) ensure that a single offset vector can characterize the modality gap, while (iii) and (iv) ensure that offset correction preserves semantic relationships.

\paragraph{Observations.}
Table~\ref{tab:geometry} demonstrates that all encoders satisfy the requirements for offset-based alignment in expectation. Gap consistency exceeds 0.96 for five of six modalities, and mean orthogonality remains near zero across all encoders, indicating that properties (ii) and (iv) hold on average. However, the degree to which these properties hold at the instance level varies substantially. We find that the variance of property (iv), gap-content orthogonality, serves as a particularly informative predictor of downstream performance. To quantify this relationship, we analyze performance preservation at the individual task level, treating each evaluation metric as a separate observation. This yields 33 data points spanning retrieval benchmarks such as AudioCaps R@1, COCO R@5, and DrugBank MRR, as well as classification benchmarks such as ModelNet40 Top-1 and ESC-50 accuracy. Since tasks evaluated on the same encoder share identical orthogonality variance, they appear vertically aligned in Figure~\ref{fig:correlation}. The analysis reveals a moderate negative correlation between orthogonality variance and performance preservation, with Pearson $r = -0.67$ and $p < 0.001$. Encoders exhibiting lower variance in property (iv), notably Uni3D at $\pm 0.04$ and ViCLIP at $\pm 0.06$, achieve preservation rates consistently above 80\%, whereas those with higher variance such as CLAP at $\pm 0.15$ and MoleculeSTM at $\pm 0.18$ show greater performance degradation. This pattern suggests that inconsistent orthogonality introduces variable distortions during offset correction, with fine-grained retrieval tasks affected more severely than categorical classification. We additionally note that MoleculeSTM exhibits a distinct failure mode in property (ii), as its gap consistency of only 0.78 indicates that a single offset vector inadequately characterizes the modality gap for molecular embeddings.

\definecolor{strongpos}{HTML}{C8E6C9}    
\definecolor{medpos}{HTML}{DCEDC8}       
\definecolor{lightpos}{HTML}{F1F8E9}     
\definecolor{lightneg}{HTML}{FFEBEE}     

\begin{table}[t]
\caption{\textbf{Effect of offset correction.} Modalities with strong gap consistency benefit substantially, while Molecule with weak consistency shows degradation.}
\label{tab:ablation_offset}
\centering
\small
\resizebox{\columnwidth}{!}{
\begin{tabular}{llccc}
\toprule
Modality & Benchmark & w/o offset & w/ offset & $\Delta$ \\
\midrule
3D & ModelNet40 & 4.05 & 70.86 & \cellcolor{strongpos}{$+94.30\%$} \\
3D & ScanObjectNN & 5.40 & 42.15 & \cellcolor{strongpos}{$+87.20\%$} \\
Audio & AudioCaps & 8.68 & 15.35 & \cellcolor{medpos}{$+43.50\%$} \\
Audio & Clotho & 4.77 & 7.81 & \cellcolor{medpos}{$+38.90\%$} \\
X-ray & RSNA & 31.35 & 46.59 & \cellcolor{lightpos}{$+32.70\%$} \\
Molecule & DrugBank & 36.44 & 34.75 & \cellcolor{lightneg}{$-4.60\%$} \\
\bottomrule
\end{tabular}
}
\end{table}

\subsection{How Do Design Choices Affect Performance?}
\label{sec:ablation}

\paragraph{Effect of Offset Correction.}
Table~\ref{tab:ablation_offset} examines the contribution of offset correction across modalities with varying gap consistency. For modalities satisfying strong gap consistency above 0.95, offset correction yields substantial improvements, with 3D classification increasing from 4.05\% to 70.86\% on ModelNet40 and Audio retrieval improving by 43.5\% on AudioCaps. In contrast, Molecule with gap consistency of only 0.78 exhibits a slight performance degradation of 4.6\%, indicating that unreliable offset estimation can introduce harmful distortions. These results suggest that practitioners should verify gap consistency before applying geometric alignment.

\begin{table}[t]
\caption{\textbf{Anchor space comparison.} LLM-based anchors yield stronger Text$\to$X retrieval performance, while multimodal anchors achieve comparable results on classification.}
\label{tab:ablation_anchor}
\centering
\scriptsize
\setlength{\tabcolsep}{4pt}
\begin{tabular}{l ccc ccc}
\toprule
& \multicolumn{3}{c}{Retrieval} & \multicolumn{3}{c}{Classification} \\
\cmidrule(lr){2-4} \cmidrule(lr){5-7}
Anchor & \makecell{AudioCaps\\R@1} & \makecell{Clotho\\R@1} & \makecell{DrugBank\\MRR} & \makecell{3D\\Top-1} & \makecell{Audio\\Top-1} & \makecell{X-ray\\Top-1} \\
\midrule
\multicolumn{7}{l}{\textit{Multimodal encoders}} \\
CLIP & 15.91 & 6.60 & \textbf{36.44} & 78.04 & \textbf{86.70} & \textbf{48.31} \\
LanguageBind & 14.54 & 6.93 & 29.66 & \textbf{81.12} & 74.65 & 44.99 \\
\midrule
\multicolumn{7}{l}{\textit{LLM embedding models}} \\
NV-Embed-v2 & \textbf{16.20} & 7.75 & 26.27 & 76.30 & 79.40 & 48.59 \\
Qwen3-Embed & 15.35 & \textbf{7.81} & 34.75 & 70.86 & 77.25 & 46.59 \\
\bottomrule
\end{tabular}
\end{table}

\paragraph{Anchor Space Selection.}
Table~\ref{tab:ablation_anchor} compares two categories of anchor spaces, which are LLM embeddings and multimodal encoders. The results reveal a task-dependent pattern in anchor space effectiveness. For retrieval tasks, LLM-based anchors such as NV-Embed-v2 and Qwen3-Embedding consistently outperform multimodal encoders on audio benchmarks, achieving 16.20 and 15.35 R@1 on AudioCaps compared to 14.54--15.91 for multimodal anchors. We attribute this advantage to the semantic representations acquired through large-scale language pretraining, which capture fine-grained similarity relationships required for retrieval. For classification tasks, multimodal anchors such as CLIP and LanguageBind demonstrate superior performance, with CLIP achieving 86.70 on Audio and LanguageBind achieving 81.12 on 3D modality. We attribute this advantage to the discriminative decision boundaries acquired through vision-language contrastive training. Based on these findings, we adopt Qwen3-Embedding as the default anchor space, as it provides balanced performance across retrieval and classification.

\begin{table}[t]
\caption{\textbf{Effect of training data source.} Domain-specific captions substantially outperform general-purpose text across all modalities.}
\label{tab:ablation_data}
\centering
\scriptsize
\resizebox{\columnwidth}{!}{
\setlength{\tabcolsep}{4pt}
\begin{tabular}{l cccc}
\toprule
Training Data & \makecell{Audio\\R@1} & \makecell{3D\\Top-1} & \makecell{X-ray\\Top-1} & \makecell{Mol.\\MRR} \\
\midrule
all-NLI & 6.36 & 12.10 & 22.48 & 16.10 \\
Domain captions & 15.35 & 70.86 & 46.59 & 34.75 \\
\midrule
Improvement & $+141\%$ & $+485\%$ & $+107\%$ & $+116\%$ \\
\bottomrule
\end{tabular}
}
\end{table}

\paragraph{Training Data Source.}
To validate whether general-purpose text corpora that have never been associated with any target modality can enable cross-modal transfer, we train projection networks using all-NLI, a corpus combining MNLI~\citep{williams2018broad} and SNLI~\citep{bowman2015large} with 100K sentence pairs, following the previous work of text-only training~\citep{xiao2025scaling}. Table~\ref{tab:ablation_data} presents the results. Training on all-NLI yields substantially lower performance compared to domain-specific captions across all modalities, with the degradation being most pronounced for 3D, which drops from 70.86\% to 12.10\%. This performance gap reflects the distributional mismatch between general linguistic expressions and the specialized vocabularies characteristic of each modality domain. Nevertheless, the non-trivial cross-modal transfer achieved with general-purpose text validates our text-only training paradigm. A systematic analysis of how distributional characteristics such as domain coverage and vocabulary specificity influence alignment quality presents a promising avenue for developing more refined data selection strategies.

\section{Related Work}

\paragraph{Modality Expansion.}
Contrastive learning has enabled effective multimodal alignment by projecting different modalities into shared semantic spaces \citep{radford2021learning, jia2021scaling}. Subsequent work extends this paradigm to multiple modalities through central hubs: ImageBind \citep{girdhar2023imagebind} uses images as the anchor modality, while LanguageBind \citep{zhu2023languagebind} leverages text for broader semantic coverage. To reduce computational costs, recent methods connect frozen pretrained encoders through lightweight projectors. C-MCR \citep{wang2023connecting} and Ex-MCR \citep{zhang2024extending} learn adapters between encoder pairs, while FreeBind \citep{wang2024freebind} and OmniBind \citep{wang2024omnibind} ensemble multiple encoders per modality. However, all these approaches require instance-level paired supervision during training, which becomes prohibitive in specialized domains where paired data is scarce. \ours\ eliminates this requirement through text-only training of projection networks.

\paragraph{Modality Gap Analysis.}
The modality gap—a systematic offset between text and non-text embeddings in contrastive models—was first identified by \citet{liang2022mind}, who characterized its geometric structure in CLIP. Subsequent work has sought to understand this phenomenon from multiple perspectives, including linear separability analysis~\citep{shi2023towards}, double-ellipsoid geometry~\citep{levi2024double}, and the distinction between modality-specific and contrastive components~\citep{fahim2024s}. Building on these insights, several methods exploit the gap for downstream applications such as vision model diagnosis~\citep{zhang2023diagnosing} and cross-modal transfer via zero-centering~\citep{zhang2024connect}. More recent efforts focus on mitigating the gap through learnable correction models~\citep{park2024bridging, eslami2024mitigate}, embedding standardization~\citep{an2025i0t}, or centroid alignment for mixed-modality retrieval~\citep{li2025closing}. However, these methods operate in paired-data settings and focus on improving alignment within existing modality pairs. While prior work has primarily analyzed the gap in vision-language models, we demonstrate that this geometric property generalizes across six diverse modalities—including 3D, X-ray, and molecules—and can be exploited for modality expansion without paired supervision.

\paragraph{LLM-Anchored Multimodal Learning.}
Recent work leverages LLMs as semantic anchors for multimodal alignment, exploiting their broad semantic coverage and contextual understanding acquired through large-scale language pretraining. Generative approaches integrate LLMs with multimodal encoders for instruction tuning~\citep{han2023imagebind} and unified cross-modal generation~\citep{han2024onellm}. Representation-focused methods include UniBind~\citep{lyu2024unibind}, which creates LLM-augmented unified spaces, and LLM2CLIP~\citep{huang2024llm2clip}, which enhances dense caption understanding through large-scale paired training on tens of millions of image-caption pairs. More recently, E5-V~\citep{jiang2024e5} and LCO-Emb~\citep{xiao2025scaling} show that text-only contrastive learning can enhance MLLM embedding quality without multimodal training data. However, these methods operate within unified MLLM architectures where cross-modal alignment is implicitly established during generative pretraining. In contrast, \ours\ addresses the alignment of independently trained contrastive encoders with architecturally heterogeneous embedding spaces, enabling expansion to specialized domains such as 3D, X-ray, and molecules without requiring a shared backbone.

\section{Conclusion}
We introduced \ours, a framework that leverages the consistent modality gap in pretrained contrastive encoders to enable text-only modality expansion. By projecting diverse modalities into LLM embedding space as a unified anchor, our approach preserves substantial performance of pretrained encoders across six modalities using only text descriptions for projection learning. Compared to existing paired-data methods, \ours\ reduces data requirements by over 95\% while eliminating the need for paired multimodal supervision. Furthermore, the framework enables emergent cross-modal retrieval between modality pairs never seen during training (\eg, audio-to-image, 3D-to-image), demonstrating that text-anchored alignment can establish implicit correspondences across arbitrary modalities without direct cross-modal pairing. These results suggest that text-only training offers a practical pathway for integrating specialized modalities—such as medical imaging and molecular structures—into unified multimodal systems without the prohibitive cost of expert annotation. 

\newpage



\section*{Impact Statement}

This work aims to reduce the data annotation burden in multimodal learning, potentially democratizing access to multimodal AI systems in specialized domains such as medical imaging and molecular analysis where paired supervision is prohibitively expensive. While this could accelerate beneficial applications in healthcare and drug discovery, practitioners should exercise appropriate caution when deploying such models in safety-critical settings, ensuring thorough validation before clinical or real-world use. We do not foresee immediate negative societal consequences beyond those common to advances in representation learning.

\bibliography{reference}
\bibliographystyle{icml2026}

\newpage

\appendix
\onecolumn

\section{Semantic Textual Similarity Benchmark Analysis}
\label{app:sts}

To validate our choice of LLM embeddings as the semantic anchor space, we conduct comprehensive evaluation on the Semantic Textual Similarity (STS) benchmark suite. Table~\ref{tab:sts} presents Spearman correlation scores across six STS tasks (STS12--16 and STSBenchmark) comparing multimodal encoders (CLIP~\citep{radford2021learning}, LanguageBind~\citep{zhu2023languagebind}) against LLM embedding models (NV-Embed-v2~\citep{lee2024nv}, Qwen3-Embedding variants~\citep{zhang2025qwen3}).

\begin{table}[h]
\caption{\textbf{Semantic Textual Similarity (STS) benchmark performance.} Spearman correlation ($\rho$) scores across six STS tasks are reported, comparing multimodal encoders and LLM embedding models.}
\label{tab:sts}
\centering
\small
\setlength{\tabcolsep}{4pt}
\begin{tabular}{l cccccc c}
\toprule
\multirow{2}{*}{Model} & \multicolumn{6}{c}{STS Tasks (Spearman $\rho$)} & \multirow{2}{*}{Avg.} \\
\cmidrule(lr){2-7}
& STS12 & STS13 & STS14 & STS15 & STS16 & STSBench & \\
\midrule
\multicolumn{8}{l}{\textit{Multimodal Encoders}} \\
CLIP & 61.87 & 63.83 & 62.09 & 76.82 & 72.89 & 72.26 & 68.29 \\
LanguageBind & 63.12 & 67.46 & 63.27 & 73.82 & 73.73 & 71.60 & 68.83 \\
\midrule
\multicolumn{8}{l}{\textit{LLM Embedding Models}} \\
NV-Embed-v2 & 77.89 & 88.30 & 84.30 & 89.04 & 86.77 & 88.41 & 85.79 \\
Qwen3-Embed-0.6B & 79.35 & 87.31 & 79.81 & 87.28 & 87.07 & 86.51 & 84.56 \\
Qwen3-Embed-4B & 84.31 & 93.20 & 88.61 & 92.31 & 92.07 & 91.92 & 90.40 \\
\bottomrule
\end{tabular}
\end{table}

\section{Extended Experimental Results}
\label{app:extended_results}

This appendix provides comprehensive experimental results that supplement the main findings in Section~\ref{sec:experiments}. We report detailed metrics across all evaluation benchmarks to enable thorough comparison and reproducibility.

\subsection{Evaluation Benchmarks}
\label{app:benchmarks}

Table~\ref{tab:benchmarks_full} provides a complete overview of all evaluation benchmarks. In the main text (Table~\ref{tab:main}), we report representative benchmarks per modality for clarity: Flickr30k for image, MSVD for video, AudioCaps for audio, and DrugBank for molecule retrieval; ESC-50, ModelNet40, ScanObjectNN, and RSNA for classification.

\begin{table}[h]
\centering
\caption{\textbf{Complete evaluation benchmarks.} All datasets used for retrieval and classification tasks are organized by task type and target modality.}
\label{tab:benchmarks_full}
\small
\begin{tabular}{lll}
\toprule
Task & Modality & Datasets \\
\midrule
\multirow{4}{*}{Text$\to$X Retrieval} 
    & Image & COCO~\citep{lin2014microsoft}, Flickr30k~\citep{plummer2015flickr30k} \\
    & Video & MSRVTT~\citep{xu2016msr}, MSVD~\citep{chen2011collecting}, DiDeMo~\citep{anne2017localizing} \\
    & Audio & AudioCaps~\citep{kim2019audiocaps}, Clotho~\citep{drossos2020clotho} \\
    & Molecule & DrugBank~\citep{knox2024drugbank} \\
\midrule
\multirow{2}{*}{Emergent X$\to$X} 
    & Audio$\to$Image & FlickrNet~\citep{senocak2018learning} \\
    & 3D$\to$Image & Objaverse~\citep{deitke2023objaverse} \\
\midrule
\multirow{3}{*}{Zero-shot Cls.} 
    & 3D & ModelNet40~\citep{sun2022benchmarking}, ScanObjectNN~\citep{uy2019revisiting} \\
    & Audio & AudioSet~\citep{gemmeke2017audio}, ESC-50~\citep{piczak2015esc} \\
    & X-ray & RSNA~\citep{rsna} \\
\bottomrule
\end{tabular}
\end{table}

\subsection{Detailed Cross-Modal Retrieval Performance}
\label{app:retrieval_detail}

Table~\ref{tab:main} in the main text reports representative R@1 metrics for brevity. Here, we provide complete retrieval results including R@5 metrics, which capture whether relevant items appear within the top-5 retrieved candidates. This relaxed criterion is particularly informative for assessing approximate semantic alignment quality.

\begin{table}[h]
\caption{\textbf{Detailed Text$\to$Image and Text$\to$Video retrieval performance.} R@1 and R@5 metrics are reported. PPR denotes Performance Preservation Ratio (\%) relative to pretrained encoders (CLIP for Image, ViCLIP for Video).}
\label{tab:retrieval_detail}
\centering
\small
\setlength{\tabcolsep}{3pt}
\begin{tabular}{l cccc cccc}
\toprule
& \multicolumn{4}{c}{Image Retrieval} & \multicolumn{4}{c}{Video Retrieval} \\
\cmidrule(lr){2-5} \cmidrule(lr){6-9}
& \multicolumn{2}{c}{COCO} & \multicolumn{2}{c}{Flickr30k} & \multicolumn{2}{c}{MSRVTT} & \multicolumn{2}{c}{MSVD} \\
& R@1 & R@5 & R@1 & R@5 & R@1 & R@5 & R@1 & R@5 \\
\midrule
Pretrained & 48.29 & 72.51 & 77.70 & 94.16 & 37.00 & 63.70 & 51.06 & 78.29 \\
TextME & 28.63 & 54.81 & 51.66 & 77.90 & 26.40 & 50.50 & 45.82 & 77.01 \\
\midrule
PPR (\%) & 59.3 & 75.6 & 66.5 & 82.7 & 71.4 & 79.3 & 89.7 & 98.4 \\
\bottomrule
\end{tabular}
\end{table}

As shown in Table~\ref{tab:retrieval_detail}, TextME consistently achieves higher PPR on R@5 compared to R@1 across all benchmarks (e.g., 75.6\% vs.\ 59.3\% on COCO, 98.4\% vs.\ 89.7\% on MSVD). This pattern indicates that text-only training effectively preserves coarse-grained semantic structure, with the performance gap primarily arising from fine-grained ranking precision. Notably, on MSVD, TextME achieves near-perfect R@5 preservation (98.4\%), suggesting that the learned projections successfully capture the underlying semantic relationships for video-text alignment.

\begin{table}[h]
\caption{\textbf{Detailed Text$\to$Audio and Text$\to$Molecule retrieval performance.} R@1 and R@5 metrics are reported for audio, and MRR@10 and MRR@20 for molecule retrieval. PPR denotes Performance Preservation Ratio (\%) relative to pretrained encoders (CLAP for Audio, MoleculeSTM for Molecule).}
\label{tab:retrieval_detail_audio}
\centering
\small
\setlength{\tabcolsep}{3pt}
\begin{tabular}{l cccc cc}
\toprule
& \multicolumn{4}{c}{Audio Retrieval} & \multicolumn{2}{c}{Molecule Retrieval} \\
\cmidrule(lr){2-5} \cmidrule(lr){6-7}
& \multicolumn{2}{c}{AudioCaps} & \multicolumn{2}{c}{Clotho} & \multicolumn{2}{c}{DrugBank} \\
& R@1 & R@5 & R@1 & R@5 & MRR@10 & MRR@20 \\
\midrule
Pretrained & 22.47 & 54.43 & 16.90 & 39.75 & 79.19 & 69.17 \\
TextME & 15.35 & 43.88 & 7.81 & 23.81 & 34.75 & 27.97 \\
\midrule
PPR (\%) & 68.3 & 80.6 & 46.2 & 59.9 & 43.9 & 40.4 \\
\bottomrule
\end{tabular}
\end{table}

The audio retrieval results exhibit a similar trend, with R@5 preservation consistently exceeding R@1. However, the molecule retrieval task shows lower overall preservation rates, which we attribute to the highly specialized vocabulary in chemical descriptions that differs substantially from the general text distributions used in LLM pretraining.

\subsection{Detailed Zero-shot Classification Results}
\label{app:classification}

Table~\ref{tab:cls_detail} provides complete zero-shot classification results including Top-5 accuracy.

\begin{table}[h]
\centering
\caption{\textbf{Detailed zero-shot classification performance.} Top-1 and Top-5 accuracy are reported across audio (ESC-50) and 3D (ModelNet40, ScanObjectNN) modalities, along with mAP for AudioSet. PPR denotes Performance Preservation Ratio (\%) relative to pretrained encoders.}
\label{tab:cls_detail}
\small
\begin{tabular}{lccccccc}
\toprule
& AudioSet & \multicolumn{2}{c}{ESC-50} & \multicolumn{2}{c}{ModelNet40} & \multicolumn{2}{c}{ScanObjectNN} \\
Method & mAP & Top-1 & Top-5 & Top-1 & Top-5 & Top-1 & Top-5 \\
\midrule
Pretrained & 9.32 & 85.20 & 97.70 & 67.75 & 90.07 & 42.21 & 77.23 \\
LanguageBind & 18.33 & 94.00 & 99.70 & -- & -- & -- & -- \\
Ex-MCR & 6.67 & 71.20 & 96.80 & 66.53 & 93.60 & 40.31 & 77.20 \\
\midrule
Na\"ive & 1.14 & 2.90 & 8.45 & 0.81 & 8.95 & 3.32 & 30.52 \\
COX$^\dagger$ & 1.26 & 2.00 & 10.00 & 4.05 & 13.70 & 2.84 & 26.68 \\
TextME & 5.80 & 77.25 & 96.85 & 70.86 & 92.14 & 42.15 & 77.89 \\
\midrule
PPR (\%) & 62.2 & 90.7 & 99.1 & 104.6 & 102.3 & 99.9 & 100.9 \\
\bottomrule
\end{tabular}
\end{table}

Notably, TextME achieves PPR exceeding 100\% on 3D classification benchmarks (ModelNet40: 104.6\%, ScanObjectNN: 99.9\%), demonstrating that text-only training can sometimes improve upon pretrained encoder performance. Top-5 preservation consistently exceeds Top-1, indicating that approximate categorical boundaries are well-preserved even when precise rankings differ.

\section{Implementation Details}
\label{app:implementation}

\subsection{Model Architecture}

Each projection network $P_m$ is implemented as a 2-layer MLP with GeLU activation:
\begin{equation}
P_m(x) = W_2 \cdot \text{GeLU}(W_1 \cdot x + b_1) + b_2
\end{equation}
where the hidden dimension matches the source encoder's embedding dimension $d_m$, and the output dimension is fixed at $d_h = 2560$ to match Qwen3-Embedding. Total trainable parameters per modality: $\sim$10M.

\subsection{Training Configuration}

\begin{table}[h]
\centering
\caption{\textbf{Training hyperparameters.} All projection networks are trained with the same configuration across modalities.}
\small
\label{tab:hyperparams}
\begin{tabular}{ll}
\toprule
Hyperparameter & Value \\
\midrule
Batch size & 512 \\
Optimizer & AdamW ($\beta_1=0.9$, $\beta_2=0.999$) \\
Weight decay & 0.01 \\
Learning rate & $5 \times 10^{-4}$ \\
LR schedule & Cosine annealing \\
Training epochs & 50 \\
Temperature $\tau$ & 0.07 \\
Hard negative range & $[0.1 \cdot s_i, 0.9 \cdot s_i]$ \\
Precision & fp16 \\
\bottomrule
\end{tabular}
\end{table}

\subsection{Offset Computation}

For each modality, we compute centroids using 5,000 randomly sampled text-modal pairs. Table~\ref{tab:offset_data} summarizes the datasets used for offset estimation across all evaluated encoders.

\begin{table}[h]
\centering
\caption{\textbf{Datasets used for offset computation.} For each encoder, centroids are estimated using 5,000 randomly sampled text-modal pairs from the listed datasets.}
\small
\label{tab:offset_data}
\begin{tabular}{llll}
\toprule
Encoder & Modality & Offset Dataset & Domain \\
\midrule
CLIP & Image & COCO~\citep{lin2014microsoft} & Natural images \\
ViCLIP & Video & MSRVTT~\citep{xu2016msr} & Web videos \\
CLAP & Audio & AudioCaps~\citep{kim2019audiocaps} & Audio events \\
Uni3D & 3D & Objaverse~\citep{deitke2023objaverse} & Synthetic objects \\
CXR-CLIP & X-ray & CheXpert~\citep{irvin2019chexpert} & Medical imaging \\
MoleculeSTM & Molecule & PubChem~\citep{kim2025pubchem} & Chemical compounds \\
LanguageBind & Text & COCO~\citep{lin2014microsoft} & Natural images \\
\bottomrule
\end{tabular}
\end{table}

Text inputs are tokenized with a maximum sequence length of 77 tokens. Offsets are pre-computed once and remain fixed throughout training.

We note that the choice of dataset for offset computation may influence both the estimated gap properties and downstream performance. Since the modality gap is computed as the difference between text and modal centroids, the semantic distribution of the offset dataset could affect the resulting offset vector. For instance, encoders whose offset datasets closely match the evaluation domain may exhibit more favorable gap properties, while domain mismatch between offset computation and downstream evaluation could introduce additional variability. Although our experiments demonstrate robust performance across diverse benchmarks, a systematic investigation of how offset dataset selection affects alignment quality remains an important direction for future work.

\subsection{Computational Resources}

All experiments are conducted on a single NVIDIA A6000 GPU (48GB). Training time averages 2 hours per modality with peak memory usage of approximately 8GB.

\section{COX Baseline Implementation}
\label{app:cox}

Since the original COX~\citep{huang2025towards} codebase is not publicly available, we re-implemented the method following the paper specifications with adaptations for our zero-shot evaluation setting.

\textbf{Architecture.} We employ Vision Transformer Tiny (ViT-T/16) as the encoder backbone with 12 layers, 3 attention heads, and embedding dimension 192. Following the original design, we incorporate a Variational Information Bottleneck (VIB) layer with stochastic dimensionality reduction to 256 dimensions.

\textbf{Training Protocol.} We follow the two-stage methodology: (1) supervised pre-training on labeled target data for 10 epochs, and (2) information bottleneck fine-tuning for 50 epochs. We use batch size 256, Adam optimizer with learning rate $1 \times 10^{-3}$ and weight decay $1 \times 10^{-5}$.

\textbf{Key Difference from TextME.} COX requires labeled target modality data ($\sim$10K samples), trains encoders from scratch ($>$300M parameters), and demands architectural alignment between source and target encoders. In contrast, TextME leverages pre-trained encoders with only text descriptions, requires merely $\sim$10M trainable parameters, and imposes no architectural constraints.

\section{Additional Ablation Studies}
\label{app:ablation}

\subsection{Sample Size for Offset Estimation}

We investigate sensitivity to the number of samples used for computing centering offsets.

\begin{table}[h]
\centering
\caption{\textbf{Impact of sample size for offset estimation.} Performance remains stable for $N \geq 1{,}000$.}
\small
\label{tab:sample_size}
\begin{tabular}{lcccc}
\toprule
\# Samples & Audio R@1 & 3D Top-1 & Mol. MRR & Rel. Perf. \\
\midrule
100 & 14.91 & 70.66 & 34.75 & 90\% \\
500 & 14.77 & 70.58 & 33.05 & 95\% \\
1,000 & 14.89 & 70.62 & 36.44 & 97\% \\
5,000 (default) & 15.35 & 70.86 & 34.75 & 100\% \\
10,000 & 14.95 & 70.58 & 32.20 & 100\% \\
\bottomrule
\end{tabular}
\end{table}

Results demonstrate that offset estimation is robust to sample size, with performance plateauing between 1,000--10,000 samples. Even with only 100 samples, the method achieves 90\% of default performance, validating the efficiency of our approach.

\subsection{Offset Noise Sensitivity}

To assess robustness to offset estimation errors, we perturb the pre-computed offset $\Delta$ with additive Gaussian noise: $\Delta' = \Delta + \mathcal{N}(0, \sigma^2 I)$.

\begin{table}[h]
\centering
\caption{\textbf{Offset noise sensitivity.} Text$\to$Audio retrieval (R@1), 3D classification (Top-1), and Text$\to$Molecule retrieval (MRR) are reported. Performance degrades gracefully for $\sigma < 0.05$.}
\small
\label{tab:noise}
\begin{tabular}{lccc}
\toprule
Noise $\sigma$ & Audio R@1 & 3D Top-1 & Mol. MRR \\
\midrule
0.000 & 14.95 & 70.46 & 27.97 \\
0.001 & 14.95 & 70.30 & 24.58 \\
0.01 & 15.04 & 67.50 & 22.88 \\
0.05 & 14.93 & 34.32 & 17.80 \\
0.10 & 14.25 & 14.91 & 11.02 \\
\bottomrule
\end{tabular}
\end{table}

Audio demonstrates remarkable stability, maintaining near-baseline performance even at $\sigma = 0.10$. In contrast, 3D and Molecule show sharper degradation at $\sigma \geq 0.05$, indicating these modalities require more precise offset estimation. For practical deployment, 5,000 samples provide sufficient precision with empirical standard error well below $\sigma = 0.01$.

\end{document}